\documentclass[10pt,twocolumn,letterpaper]{article}

\usepackage{cvpr}
\usepackage{times}
\usepackage{epsfig}
\usepackage{graphicx}
\usepackage{amsmath}
\usepackage{amssymb}
\usepackage{epstopdf}

\usepackage{booktabs}
\usepackage{tabularx}
\usepackage{multirow}

\usepackage[pagebackref=true,breaklinks=true,letterpaper=true,colorlinks,bookmarks=false]{hyperref}

 \cvprfinalcopy 

\DeclareMathOperator*{\argmin}{arg\!\min}

\def\cvprPaperID{3489} 

\ifcvprfinal\fi
\begin{document}

\title{Guide Me: Interacting with Deep Networks}

\author{
	Christian Rupprecht$^{1,2,\textbf{*}}$, \hspace{1mm} 
	Iro Laina$^{1,\textbf{*}}$, \hspace{1mm}  
	Nassir Navab$^{1}$, \hspace{1mm}  
	Gregory D. Hager$^{2,1}$, \hspace{1mm}  
	Federico Tombari$^{1}$ 
	\\ \\
	$^{\textbf{*}}$ equal contribution \\
	$^{1}$ Technische Universit{\"a}t M{\"u}nchen, Munich, Germany \\ 
	$^{2}$ Johns Hopkins University, Baltimore, MD, USA}

\maketitle
\begin{abstract}
	Interaction and collaboration between humans and intelligent machines has become increasingly important as machine learning methods move into real-world applications that involve end users. 
	While much prior work lies at the intersection of natural language and vision, such as image captioning or image generation from text descriptions, less focus has been placed on the use of language to guide or improve the performance of a learned visual processing algorithm.	
	In this paper, we explore methods to flexibly guide a trained convolutional neural network through user input to improve its performance during inference. We do so by inserting a layer that acts as a spatio-semantic guide into the network. This guide is trained to modify the network's activations, either directly via an energy minimization scheme or indirectly through a recurrent model that translates human language queries to interaction weights. Learning the verbal interaction is fully automatic and does not require manual text annotations. We evaluate the method on two datasets, showing that guiding a pre-trained network can improve performance, and provide extensive insights into the interaction between the guide and the CNN. 
\end{abstract}

\section{Introduction}
\begin{figure*}[ht]
    \centering
    \includegraphics[trim = 180mm 100mm 180mm 110mm, clip, width=0.9\linewidth]{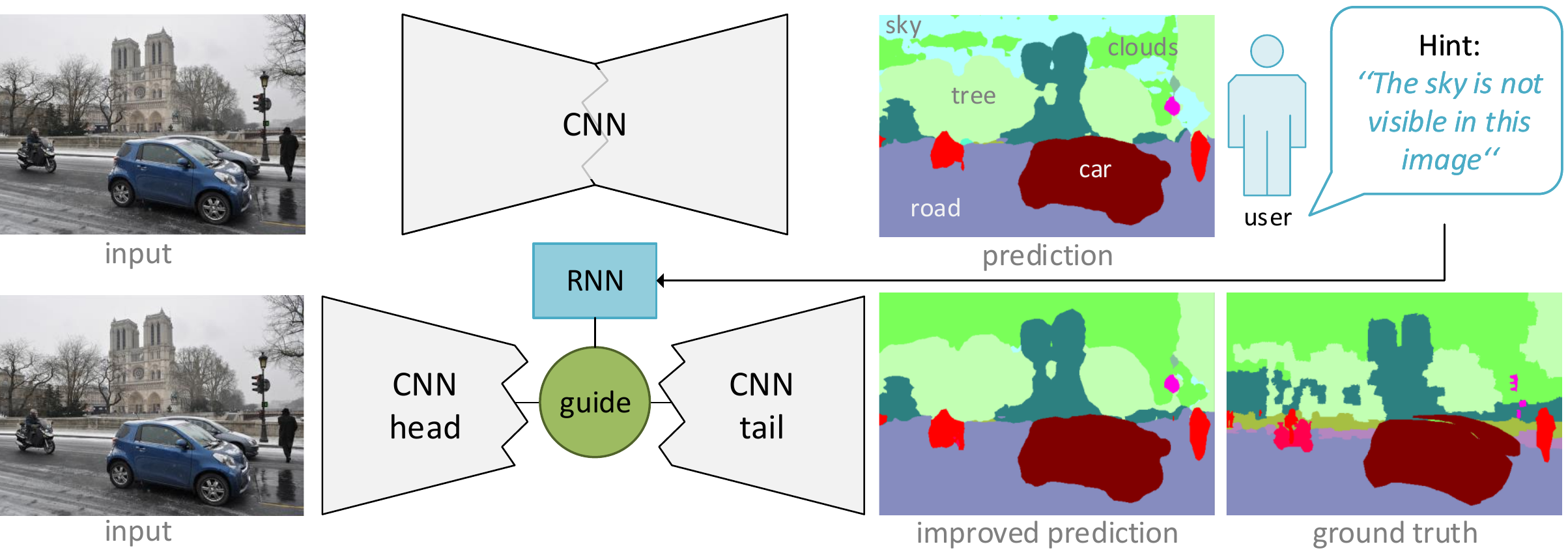}
    \caption{\textbf{Overview.} We introduce a system that is able to refine predictions of a CNN by injecting a guiding block into the network. The guiding can be performed using natural language through an RNN to process the text. In this example, the original network had difficulties to differentiate between the \texttt{sky} and the \texttt{cloud} classes. The user indicates that there is no sky and the prediction is updated, without any CNN weight updates and thus without additional training.}
    \label{fig:teaser}
\end{figure*}

Convolutional neural networks (CNNs) continue to grow in their breadth of application and in their performance on challenging computer vision tasks, such as image classification, semantic segmentation, object detection, depth prediction and human pose estimation. 
To date, the majority of the techniques proposed for these applications train specific network architectures once and subsequently deploy them as static components inside an algorithm.
However, it is unlikely that any static network will be perfect at the task it was designed for.
If the deployed CNN were adaptable to feedback or specifications provided by a human user \emph{online}, this interaction would hold the potential to improve the model's performance and benefit real-world applications. 

For example, in photo editing, when a CNN is used to segment the foreground of an image from the background, the user might notice that the network has made a mistake. Instead of manually repairing the segmentation output or developing a post-processing algorithm based on some heuristics, a simpler and more effective way would be for the user to interact directly with the network through a directed hint, \eg pointing out that ``\textit{the child on the bottom left of the image is in the foreground, not in the background}''. 
The user that was previously presented with a fixed, black-box prediction is now able to influence and alter the outcome according to his needs.
This property is particularly useful in high risk domains such as medical image analysis and computer-assisted diagnosis, where fully automated segmentation is not always robust in clinical applications and the experience of trained practitioners matters.
Another relevant example is speeding up labor-intensive and repetitive labeling tasks, such as those needed to create datasets for semantic segmentation, especially those for which annotations are scarce and expensive to obtain.  

We propose a novel idea to allow user-network feedback-based interaction that aims at improving the performance of a pre-trained CNN at \emph{test} time. 
The core idea is the definition of a spatio-semantic guiding mechanism that translates user feedback into changes in the internal activations of the network, thus acting as a means of re-thinking the inference process. 
The user input is modeled via a language-based approach, that enables interaction with a trained model in the form of a dialog. The user receives a first estimate, inputs a text query and the network replies with an updated prediction.
Most previous interactive approaches place the user on the input/data side which means user input is required for the method to operate. In contrast, in our method, the user's input is optional and modifies the network, this means that the network does not depend on human interaction but can be adjusted by it.

We showcase this interactive module on the task of semantic image segmentation. One advantage of our method is that it does not depend on any explicit annotation for text-region correspondences.  
Yet, results indicate that the module can successfully transfer semantic information from the natural language domain to the visual domain, such that the network eventually produces a more accurate segmentation. 
As a side effect, this provides interesting insights into how CNNs structure their inference with respect to natural categories, providing an avenue for exploring the relationship between language and imagery.

\section{Related Work}
\paragraph{Interaction with neural networks}
Human-machine interaction is an extensively researched field. In \cite{branson2014ignorant} the user and an algorithm work together to solve fine grained recognition tasks, leveraging analytic strengths of the system and visual recognition power of the human. Prior to deep learning, several systems have been proposed for semi-automatic segmentation, that allowed the user to interfere with the result or to provide hints to the system via seed points \cite{boykov2001interactive,price2010geodesic}, bounding boxes \cite{grady2011segmentation,lempitsky2009image,rother2004grabcut}, contours or scribbles \cite{batra2010icoseg,grady2006random,xu1998snakes}, eye gaze \cite{hebbalaguppe2013interaction} or in the form of binary yes/no answers to a set of questions \cite{chen2016interactive,rupprecht2015image}.

Most deep learning based segmentation methods, however, do not have an interface for human input during inference. The model and thus the attainable performance is fixed after the training phase. Directly integrating a human into a training loop with thousands of images is challenging.
Nonetheless, some methods towards interactive deep learning have been proposed, such as weakly-supervised semantic segmentation from scribbles \cite{lin2016scribblesup}, user-provided clicks and Euclidean distance maps \cite{xu2016deep} or bounding boxes used as region initialization \cite{dai2015boxsup, rajchl2017deepcut}. Additionally, a method for sparse, user-guided colorization of grayscale images is proposed in \cite{zhang2017real}.
In the field of medical imaging, \cite{amrehn2017ui} proposes to interactively improve segmentation by updating a seed-map given by the user and \cite{wang2017deepigeos} uses a second network operating both on the previous prediction and human feedback. 

In our system, we integrate online interaction into the training by substituting the human input with an algorithm that dynamically generates hints from different modalities based on previous predictions. The CNN is already trained and only asks for the user's directions for the purpose of conditioned (on-demand) adjustments of an initial estimate.

\vspace{-1em}
\paragraph{The intersection of vision and language}
To enable user interaction in a natural and intuitive way, we propose a novel idea that lies in the joint domain of natural language and vision.
A relevant line of work in this field is Visual Question Answering (VQA). A question is posed and the answer is based on the image context \cite{agrawal2017vqa,malinowski2015ask,zhou2015simple}. 
Specialized systems for VQA ground the question in the input image and focus on the relevant parts to answer complex queries with text responses \cite{andreas2016learning, andreas2016neural, hu2017learning, johnson2017inferring, lu2016hierarchical, malinowski2017ask}. 
Other examples include generation of referring expressions \cite{KazemzadehOrdonezMattenBergEMNLP14,luo2017comprehension,mao2016generation,yu2016modeling}, segmentation or object retrieval from referring expressions \cite{de2017guesswhat, hu2016natural, hu2016segmentation}, image captioning \cite{dai2017towards, donahue2015long, Johnson_2016_CVPR, karpathy2015deep, mao2014deep, vinyals2015show} and visual dialog \cite{das2016visual}. 
Most works focus on the combination of CNN and RNN models, often building attention mechanisms \cite{anderson2017bottom, lu2016knowing, xu2016ask, xu2015show}. 
Most related to our approach is the recent method from \cite{de2017modulating,perez2017film} that proposes the use of a conditional batch normalization layer \cite{ioffe2015batch} and feature-wise adjustment for visual reasoning.

A key distinction between our approach and most of the summarized literature is that our system's output is visual and not textual, \ie it is neither an answer nor an image caption. The output of the interactive CNN is in the same domain as the initial one. Another major difference is that we do not rely on vision-text correspondences with paired questions-answers or captions; user interaction is simulated via textual expressions that we generate automatically.

\vspace{-1em}
\paragraph{Semantic segmentation}
In this paper, we focus on the application of semantic image segmentation, which is widely studied in the computer vision literature and significantly boosted by the success of deep learning methods \cite{badrinarayanan2015segnet, chen2016deeplab, jegou2017one, lin2016refinenet, long2015fully, pohlen2016full}.
Our goal is to deploy out-of-the-box, state-of-the-art models \cite{chen2016deeplab, long2015fully} as estimators, that are then guided by our module to improve their former predictions with the help of a human user (or any given priors as hints). 
\section{Methods}

In this section, we describe how our interaction module is integrated into a \emph{fixed} CNN following two different approaches: guiding with user clicks and back-propagation (Section \ref{method:backprop}) or natural language inputs (Section \ref{method:text}). A general overview for the latter case is shown in Figure \ref{fig:teaser}. 

We first define the main elements of our framework, which we refer to throughout this paper. 
The module we insert into the CNN is called the \textit{guide}. The guide interacts with the \textit{guided CNN} through a \textit{guiding block}, which is built to adjust activation maps of the CNN to improve the final prediction for the given input. The guided CNN is thus split into two parts: the \textit{head}, which processes the input until it reaches the guiding block, and the \textit{tail}, that is the rest of the guided network up to the final prediction layer. 
More formally, by decomposing the network into a head $h$ and a tail function $t$, the output prediction $\tilde{y}$ given input $x$ can be written as $t(h(x)) = \tilde{y}~$. 
We refer to the  information that the guide uses to modify the guided network as the \textit{hint}.

The split position is chosen manually. However, a reasonable choice is the (spatially) smallest encoding that the network produces, as this layer likely contains the most high-level information in a compact representation.We validate the choice of layers in the Section \ref{sec:guidewtext}.

\subsection{Guiding Block}
The guiding block is the integral piece of our approach, it enables feedback to flow from the guide into the guided network. Essentially, the guide must be able to modify a set of activations in the guided network. 
Since activation maps usually consist of a large number of elements (\eg $32\times32\times1024 \approx$ 1 million), element-wise control is prone to over-fitting. 
The intuition behind our guiding block is that the network encodes specific features in each channel of a given layer. Thus, a guide that has the ability to strengthen or weaken the activations per channel, can emphasize or suppress aspects of the image representation and thus adapt the prediction in a semantically meaningful way. 

The head predicts a feature representation $h(x) = A \in \mathbb{R}^{H\times W \times C}$ with width $W$, height $H$ and number of channels $C$. Then, guiding can be expressed as a per feature map multiplication with a vector $\gamma^{(s)} \in \mathbb{R}^C$ and bias $\gamma^{(b)} \in \mathbb{R}^C$,
\begin{equation}\label{eq:simpleguide}
	A'_c = (1 + \gamma^{(s)}_c)A_c + \gamma^{(b)}_c~,
\end{equation}
where $c \in [1, \ldots, C]$ indexes the channels of the feature volume $A$ and the corresponding elements of the guiding vector $\gamma = (\gamma^{(s)},\gamma^{(b)})$. Given this formulation, we are able to adjust a set of feature maps by emphasizing or hiding information per channel. Equation \eqref{eq:simpleguide} can also be interpreted as the guide predicting a residual update (similar to ResNets \cite{he2016deep}) for the activation map $A_c$. 
$\gamma$ plays the role of a filter on the feature maps. 
When $\gamma = 0$, our guiding block reproduces the input feature map and thus has no effect in guiding the network. When $\gamma^{(s)}_c = -1$, channel $c$ would become suppressed as all its units would be set to 0. Conversely, for $\gamma^{(s)}_c = 1$, the activation strength of that feature channel is doubled. Values smaller than $-1$ \textit{invert} a feature map, emphasizing aspects that would have been otherwise cut-off by the ReLU unit that typically follows the weight layer (or vice versa). 

While this approach, which is similar to the conditioning layer in \cite{de2017modulating,perez2017film}, supports per-channel updates and feature re-weighting via $\gamma$, it is not flexible enough to adjust features spatially since it modifies the whole feature map with the same weight. In other words, it is impossible for this module to encourage spatially localized changes in each feature map (``\textit{On the top left you missed ...}''). To overcome this limitation, we extend the approach to the spatial dimensions $H$ and $W$ of the feature map, \ie we introduce two additional guiding vectors $\alpha \in \mathbb{R}^H$ and $\beta \in \mathbb{R}^W$ to modify the feature map $A$ with spatial attention. In the following, we will index $A$ with $h$, $w$ and $c$ to uniquely identify a single element $A_{h,w,c} \in \mathbb{R}$ of $A$
\begin{equation}\label{eq:fullguide}
	A'_{h,w,c} = (1 + \alpha_h + \beta_w + \gamma^{(s)}_c) A_{h,w,c} + \gamma^{b}_c
\end{equation}
The overall function that the \emph{guided} network computes is thus modified to
\begin{equation}\label{eq:fullguidefunc}
	y^* = t\left[(1 \oplus \alpha \oplus \beta \oplus \gamma^{(s)}) \odot h(x) \oplus \gamma^{(b)}\right]~,
\end{equation}
where the tiling of the vectors $\alpha$, $\beta$, $\gamma$ along their appropriate dimensions is denoted with $\oplus$ and the Hadamard product with $\odot$.
This way $\alpha$ and $\beta$ have spatial influence and $\gamma$ controls the semantic adjustment. 
Guiding with Equation \ref{eq:fullguidefunc} reduces the number of parameters from $H\times W \times C$ to $H + W + C = 1088$ in the previous example, which is manageable to predict with a small guiding module.
 
Since \textit{fully convolutional} architectures are a common choice for image prediction tasks, we employ linear interpolation of $\alpha$ and $\beta$ when the feature map spatial resolution varies. This choice reflects two properties of the guiding block. First, $\alpha$ and $\beta$ do not depend on fixed $H$ and $W$. Second, one can select the granularity of the spatial resolution by changing the dimensionality of $\alpha$ and $\beta$ to match the spatial complexity of the hints that the guide follows.

We describe two fundamentally different ways to employ the guiding block. The first one -- guiding by back-propagation (Section \ref{method:backprop}) -- can be directly applied on a pre-trained CNN that is kept constant.  
The second one aims at online interaction with neural networks via user feedback. The network should be able to deal with hints from different modalities, such as natural language -- ``\textit{the dog was mistaken for a horse}''. We describe how the guiding parameters $\alpha$, $\beta$ and $\gamma$ can be predicted with an appropriate module given a hint from a different input domain in Section \ref{method:text}, which also speeds up processing.

\subsection{Guiding by Back-propagation}
\label{method:backprop}
In this setup, our goal is to optimize the guiding parameters such that the network revises its decision making process and, without further learning, improves its initial prediction for the current input.
The guiding block is placed between head and tail, and the guiding parameters are initialized to $0$. For a given sample $x$, we formulate an energy minimization task to optimize $\alpha$, $\beta$ and $\gamma$. The hint will be given as a sparse input $\hat{y}$ associated to a mask $\hat{m}$. $\hat{y}$ and $\hat{m}$ have the same dimensionality as the prediction $\tilde{y}$ and the ground truth $y$. $\hat{m}$ is a binary mask that indicates the locations where a hint (\ie prior knowledge) is given. Prior knowledge can be either directly given by the user or it could be a prior computed by another source.

In semantic segmentation, one can think of the hint as a single (or more) pixel(s) for which the user provides the true class -- ``\textit{this} [pixel] \textit{is a dog}'' as additional information. Prior to guiding, a certain loss $\mathcal{L}(t(h(x)), y)$ has been minimized during training of the network for a given task. We now optimize towards the same objective, \eg per-pixel cross entropy for segmentation, but use the mask $\hat{m}$ to only consider the influence of the hint and minimize for the guiding parameters, as opposed to the network's parameters, \ie
\begin{equation}\label{eq:guidebybackprop}
\alpha^*, \beta^*, \gamma^* = \argmin_{\alpha, \beta, \gamma}\left[ \hat{m} \odot \mathcal{L}(y^{*}, \hat{y})\right]~.
\end{equation}

In this case, we only update the guiding variables for the \emph{current} specific input $x$ and hint $\hat{y}$, whereas the network's weights are not trained further. 
The minimization finds the best parameters $\alpha^*, \beta^*, \gamma^*$ conditioned on the hint. 
The key insight is that this results in an overall adjusted prediction.

Since the guiding block and the network $t(h(x))$ are differentiable, we can minimize \eqref{eq:guidebybackprop} using standard back-propagation and gradient descent with momentum. Intuitively, the tendency of gradient descent to fall into local optima is desirable here. We are looking for the smallest possible $\alpha$, $\beta$ and $\gamma$ that brings the guided prediction closer to the hint while avoiding degenerate solutions such as predicting the whole image as the hinted class.

\subsection{Learning to Guide with Text}
\label{method:text}
While the previous idea is straightforward and simple to apply to any network, it requires the hint to be given in the same domain as the network's output. 
We now explore a more natural way of human-machine interaction, in which the user can give hints to the network in natural language and the guiding mechanism is trained to update its parameters after parsing the user's inputs. 
To the best of our knowledge, this is a topic that has not been previously studied.
\vspace{-1em}
\paragraph{Training with queries} 
Similar to prior work in related fields, we use a recurrent neural network (RNN) for processing natural language inputs. We first encode the input query using a word embedding matrix, pre-trained on a large corpus, 
to acquire a fixed-length embedding vector per word. The embedded words are fed as inputs to a Gated Recurrent Unit (GRU) \cite{cho2014properties,chung2014empirical} at each time step. We freeze and do not further train the word embedding alongside our guiding module, to retain a meaningful word representation.  
The guiding parameters $\alpha$, $\beta$ and $\gamma$ are predicted as a linear mapping from the final hidden state of the GRU. 

The language-guided module is trained as follows. We first generate an initial prediction with the fixed, task-specific CNN, without influence from the guide. We then feed prediction and ground truth into a hint generator, which produces a query (e.g.~``\textit{the sky is not visible in this image}'', thus mimicking the user. 
The query is then encoded into a representation that becomes responsible for estimating $\alpha, \beta$ and $\gamma$ that will guide the feature map using \eqref{eq:fullguidefunc} and subsequently update the prediction. The standard loss for the given task (pixel-wise cross entropy loss for semantic segmentation) is re-weighted giving positive weight (1) to the class(es) mentioned in the query to encourage changes in the prediction that coincide with the given hint. All wrongly predicted pixels are given a zero weight to prevent hints from being associated with other visual classes. Initially correct pixels are weighted $0.5$ to discourage corrupting correctly classified regions.

\begin{figure}
	\centering
	\includegraphics[trim = 35mm 70mm 100mm 40mm, clip,width=0.9\linewidth]{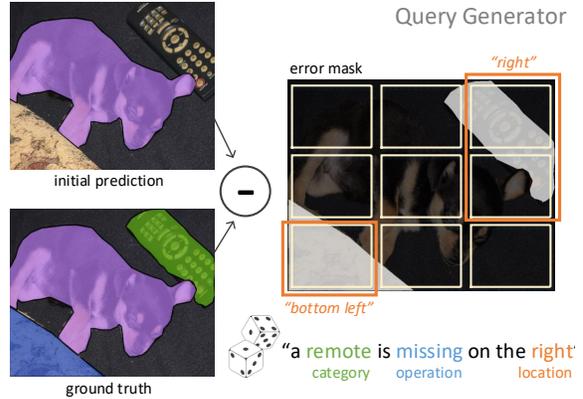}
	\caption{\textbf{Query Generator}. We illustrate the process to automatically generate queries to substitute the user during training.} \label{fig:querygen}
\end{figure}

\vspace{-1em}
\paragraph{Generating queries} 
Previous work mostly relies on human-annotated queries, which make them rich in variety; however, in this case they would limit the model to a single interaction, since new annotations cannot be recorded adaptively during training. 
Instead, our approach uses vision-only datasets and does not require visual/textual annotations, such as captions \cite{lin2014microsoft}, referring expressions \cite{hu2016segmentation, kazemzadeh2014referitgame} and region-description correspondences \cite{mao2016generation}. Our method aims at aiding the network to correct predictions with various mistakes, rather than producing a segmentation result on demand given an input expression. Therefore, it requires textual expressions that are synthesized on-the-fly from visual categories, by comparing the initial prediction and the ground truth segmentation map. 

For the generation of the queries we use a combination of functionality, semantic categories and spatial layouts (see Figure \ref{fig:querygen}). Functional categories are defined by a set of operations that can be carried out on the output to improve it, such as discovering missing semantic classes, suppressing noise or replacing wrongly predicted pixels with another class. The set we used in our experiments consists of two operations, \ie \texttt{find} to handle classes missing in the initial prediction and \texttt{remove} to correct wrongly predicted labels. 

Each query is built by its function and two placeholders, the entries of which are randomly selected at each training step from a set of plausible combinations based on prediction and the ground truth. 
We first divide the output of the network into a $N \times N$ grid. In each grid cell, we search for all erroneous classes, either missing or mistaken, while ignoring tiny spurious regions comprised of only a few pixels.
Next, we randomly sample a class from the generated list of possible choices and use its semantic name for the textual expression (\eg ``\textit{find the person}''). The sampling probability is proportional to the potential improvement in the prediction. We then track the class position in the image based on the cells where it was found. Different combinations of cells define different spatial attention areas which can be then converted into text phrases such as ``\textit{on the top left}'', ``\textit{on the bottom}'', ``\textit{in the middle}''.
 
Eventually, the proposed approach can generate textual phrases automatically and online. The guide is thus trained to understand language using vision-only annotations, \ie segmentation masks.
The guiding block is able to discover semantic and spatial correspondences between the text input and the visual feature representation.
During testing the guide can then interpret the commands of a real user.

\section{Experiments}
We evaluate our guiding framework under multiple aspects. First, we guide semantic segmentation by back-propagation. This allows us to directly evaluate the performance of the guiding and show how it can be deployed into a model without any additional training.
Second, we thoroughly investigate guiding with textual hints. 

\begin{table}
	\centering
	\begin{tabular}{l r r r r r r}
		\toprule
		\#questions & 0 & 1 & 5 & 10 & 15 & 20  \\ 
		\midrule
		\textbf{FCN-8s} & & & & & & \\
		mIoU & 62.6 & 65.3 & 73.1 & 76.9 & 77.3 & \textbf{81.0} \\  
		p.accuracy & 91.1 & 91.8 & 94.1 & 95.3 & 96.0 & \textbf{96.3} \\
		\bottomrule
	\end{tabular} 
	\vspace{0.5em}
	\caption{\textbf{Performance after a number of questions.} We guide a pre-trained FCN-8s \cite{long2015fully} on PascalVOC 2012 \textit{val} set \cite{pascal-voc-2012} directly, using back-propagation. We report the mean intersection over union (mIoU) score and pixel accuracy. Every interaction with the user improves the result.}
	\label{tab:20q}
\end{table}

\subsection{Guiding by Backpropagation} \label{sec:backprop}
We investigate the performance gain by employing our guiding block directly on a fixed, pre-trained CNN. The task is semantic segmentation on the PascalVOC 2012 dataset \cite{pascal-voc-2012}. We use a pre-trained FCN-8s network \cite{long2015fully} and insert a guiding block in the smallest encoding layer. 

A user interaction scheme similar to the 20-question game of \cite{rupprecht2015image} is set up. After an inference step, the network is allowed to ask the user for the class of a single pixel and the guiding layer updates the feature representation using \eqref{eq:guidebybackprop}.
The queried pixel is the one with the smallest posterior probability difference between the two most confident classes. This pixel has the highest interclass uncertainty, meaning that it is the most likely to flip. After each question the prediction is updated and the mean intersection over union (mIoU) is computed.

We have intentionally chosen a somewhat ``outdated'' architecture since we believe that user interaction is mostly necessary in tasks in which the performance is not close to optimal.
We list the performance after 0, 1, 5, 10, 15 and 20 questions in Table \ref{tab:20q}, where 0 denotes the initial performance of FCN-8s without guiding. Over the course of 20 interactions with the user, a significant improvement of the performance from $62.6$\% to $81.0$\% is recorded. It is noteworthy that the top entry on the PascalVOC leaderboard (DeepLab-v3 \cite{chen2017rethinking}) currently scores $86.9$\% mIoU, when trained with additional data. This demonstrates the benefit of guiding by back-propagation: it can be directly incorporated into a pre-trained CNN and, without any further training, it boosts a comparably low performance to reach the state of the art.

\subsection{Guiding with Text Inputs}\label{sec:guidewtext}
Due to the high accuracy of current methods on the PascalVOC semantic segmentation task, bringing a human into the loop to request improvements was not found to be meaningful with state-of-the-art models. We wish to evaluate our guiding module under a more challenging setting, in which even the performance of a state-of-the-art model is not satisfactory and interaction with a user can be beneficial. 

For this purpose, we have chosen to use a dataset with a limited number of images but rich categorical context. COCO-Stuff \cite{caesar2016coco} is a subset of the popular MS Common Objects in Context (COCO) dataset \cite{lin2014microsoft} and consists of 10k images from the \textit{train} 2014 set, further split into 9k training and 1k testing images. The images are labeled with pixel-level annotations of 91 ``things'' and 91 ``stuff'' classes. 

\begin{table}[!t]
	\centering
	\begin{tabular}{l | r r }
		\toprule
		guiding module & mIoU & mIoU \\ 
		&  w/ res-blk & w/o res-blk \\ 
		\midrule
		FiLM \cite{perez2017film}  & 33.08 & 33.31 \\
		ours & 33.11 & \textbf{33.56} \\
		\bottomrule
	\end{tabular} 
	\vspace{0.5em}
	\caption{\textbf{Guiding Block Variants.} We evaluate mIoU performance when guiding res4a using \texttt{find} queries, in comparison to the conditioning layer of \cite{perez2017film}.}
	\label{tab:guidemethod}
\end{table}

\begin{table}[!t]
	\centering
	\begin{tabularx}{0.9\linewidth}{lXXXX}
		\toprule
		& res3a & res4a & res5a & res5c  \\ 
		\midrule
		mIoU & 32.21 & 33.56 & 35.97 & \textbf{36.50} \\
		\bottomrule
	\end{tabularx} 
	\vspace{0.5em}
	\caption{\textbf{Location of the guiding block.} We evaluate mIoU performance when guiding different layers inside the CNN using \texttt{find} queries.} 
	\label{tab:reslocation}
\end{table}

\vspace{-1em}
\paragraph{Implementation Details.}
We first split the training set into two halves and use the first part for pre-training a DeepLab model \cite{chen2016deeplab} with a ResNet-101 \cite{he2016deep} as back-end. 
The input dimensions are $320 \times 320 \times 3$. 
On this small, challenging dataset, this model scores only 30.5\% mIoU. 
Next, we keep the weights of the semantic segmentation model fixed and only train the guiding mechanism using the remaining 4,500 images that were unseen during the pre-training phase. The guide is trained to translate embedded text queries through a recurrent model into relevant guiding parameters, as described in Section \ref{method:text}.
For the word embedding we used a pre-trained matrix based on the GloVe implementation \cite{pennington2014glove} that projects each word into a 50-dimensional vector space. 
The GRU consists of 1024 hidden units. A dense weight layer maps the last state to the vectors $\alpha$ and $\beta$, that match respectively the height and width of the succeeding activations of the semantic segmentation model, and the weights and biases that are used as the scale and offset update for each activation map. We have experimented with two ways of applying the guide. The first one alters the CNN's activations directly, therefore the weight vector size depends on the CNN layer that is being guided. The second wraps the predicted weights inside a residual block with 256 channels, as in \cite{perez2017film}. 
For the hint generation process, instead of uniquely defining an operation as ``\textit{find the \ldots}'' we randomly select from a set of variations with similar meaning such as ``\textit{the \ldots is missing}'', or ``\textit{there is a \ldots in the image}''. The grid size $N$ is set to 3, resulting in 9 cells that specify the spatial location for the query. 
All experiments are averaged over five evaluation runs to account for the randomness in the queries. 

\begin{figure*}
	\centering
	\includegraphics[clip,width=\linewidth]{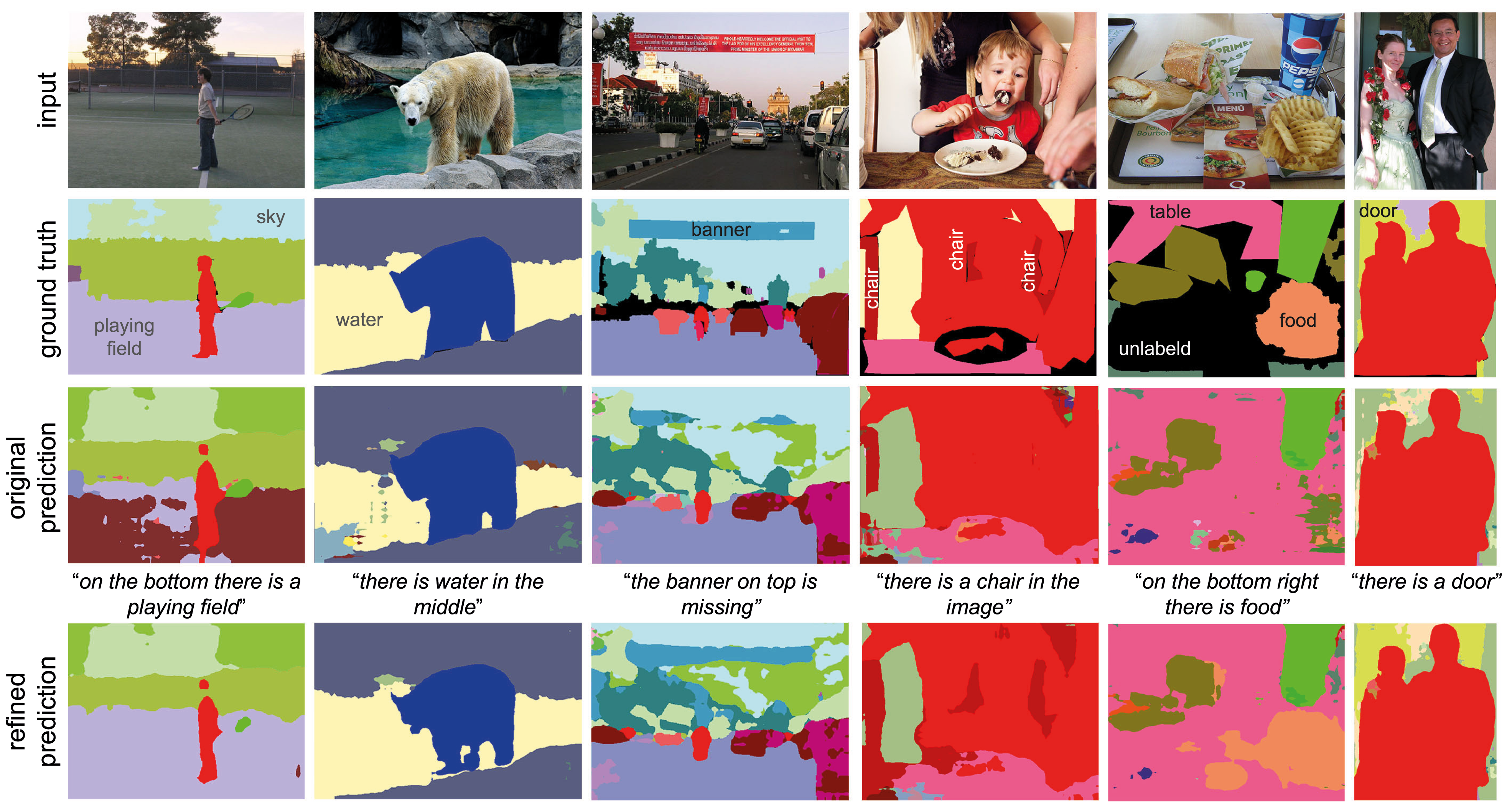}
	\caption{\textbf{Qualitative Results}. We show qualitative results using \texttt{find} hints for missing classes. In the first example, we resolve a confusion between \texttt{ground} and \texttt{playing field}. In the second example, we show that the often occurring spurious predictions can also be handled. The third column shows that the network get the hint to find the \texttt{banner}, although it bleeds slightly into the \texttt{building} below. In the fourth and the last column, classes that are heavily occluded can be discovered too after guiding. The black ground truth label stands for \texttt{unlabeled} thus any prediction is allowed there. Please see the supplementary material for additional examples.} 
	\label{fig:qualitative}
\end{figure*}

Our best guided model improves the overall score from 30.5\% to 36.5\% with a single hint. We note that training DeepLab on the full train set is only marginally better than on the half, reaching 30.8\% mIoU. Exemplary CNN predictions before and after guidance are shown in Figure \ref{fig:qualitative}. The guiding module was trained with \texttt{find} queries and does not modify the original CNN permanently, but only conditioned on the hints. We observed that our method helped resolving typical problems with the initial predictions, such as confusions between classes (columns 1, 2), partially missing objects (column 3, 4) and only partially visible objects in the background (column 6).

In the following, we compare our guiding block to the conditional batch normalization layer of \cite{perez2017film}. Then, we explore the effect of guiding location by inserting the guide at different layers of the CNN. Further, we evaluate hint complexity using different query operations and apply repeated guiding to further improve the result.
Finally, we provide some insights, by analyzing failure cases through heat map visualization and embeddings of the guiding vectors.

\begin{table}[!t]
	\centering
	\begin{tabular}{l|rr}
		\toprule 
		& \multicolumn{2}{l}{guiding location}  \\
		hint complexity  & res4a & res5a \\
		\midrule
		\texttt{remove}      &     31.53 &  32.56 \\
		\texttt{find or rmv} &     32.22 &  33.73 \\
		\texttt{find}        &     33.56 &  \textbf{35.97} \\
		\bottomrule
	\end{tabular} 
	\vspace{0.5em}
	\caption{\textbf{Complexity of Hints.} We show performance of the method using two different types of hints.}
	\label{tab:hintcomplexity}
\end{table}

\begin{table}[!t]
	\centering
	\begin{tabular}{l rrrrr}
		\toprule
		\# hints & 0 & 1 & 2 & 3 & 4 \\ 
		\midrule
		mIoU & 30.53 & 34.04 & \textbf{35.01} & 34.24 & 31.44 \\
		\bottomrule
	\end{tabular} 
	\vspace{0.5em}
	\caption{\textbf{Guiding multiple times.} We guide iteratively with multiple \texttt{find or rmv} hints. After three hints  performance decreases due to the guide over-amplifying certain features.}
	\label{tab:multiplehints}
\end{table}

\vspace{-1em}
\paragraph{Guiding Block Evaluation.}
In a set of experiments we investigate different variants of the guiding block. The performance can be seen in Table \ref{tab:guidemethod}. We analyze variants with and without an encompassing res-block around the guiding layer. We compare to the FiLM layer from Perez et al.~\cite{perez2017film}. 
The difference to our guiding block are the guiding components $\alpha$ and $\beta$, that translate location information from the text to spatial attention in image space.

\vspace{-1em}
\paragraph{Guiding Location.}
Due to the flexibility of the guiding block, it can be plugged into the network at any location. In general, in our experiments we observe that a location that is very late - close to the prediction - inside the network often results in small, local changes in the output. Moving the block earlier results in more global changes that affect a bigger region and sometimes multiple classes. When the guide is placed too early in the network the feature maps that it guides do not contain enough high-level information to guide appropriately.
This can be observed in Table \ref{tab:reslocation}, where we compute the mIoU score for guides in different locations inside the network.

\vspace{-1em}
\paragraph{Complexity of Hints.}
Automatically generating hints during training alleviates the need for manual vision-text annotations and also enables direct control of the query complexity. 
We differentiate between two distinct hints: \texttt{find} and \texttt{remove}. A \texttt{find} hint tells the network that it had missed a class: "\textit{There is a person in the top right}". \texttt{remove} is the opposite problem - the network had predicted a class that is not there or incorrect. 

In Table \ref{tab:hintcomplexity} we show the performance for the different hint types. We observe that \texttt{remove} generally yields a lower performance gain than \texttt{find}. This is explained by the fact that \texttt{remove} is a more ambiguous query than \texttt{find}. When the network is told to remove a class from the prediction it does not know what to replace it with. Training with both queries simultaneously(\texttt{find or remove}), randomly selecting one each time, achieves average performance between the two types.

\vspace{-1em}
\paragraph{Guiding multiple times.}
We have conducted an experiment, similar to the one in Section \ref{sec:backprop} and Table \ref{tab:20q}, to showcase an interesting property of the guiding module. Since it is trained to adjust the feature space in a way that improves the prediction, we hypothesize that the guided network can be guided repeatedly. 
The insight is that the guiding block will still result in a valid feature map. 
We iteratively direct the network (guided at layer res5a) to correct its mistakes via \texttt{find or rmv} queries, although it is not trained with subsequent hints, and report prediction accuracy in Table \ref{tab:multiplehints}. 
We observe that the performance has further increased with a second hint. With three or more the guide starts to over-amplify certain features, causing noise in the predictions and decreasing performance. Nonetheless, we still observe a good gain over the non-guided model. 

\subsection{Insights into the Model}
We provide further insight into learned models by examining failure cases and the learned joint embedding.

\begin{figure}[!t]
	\centering
	\includegraphics[clip,width=\linewidth]{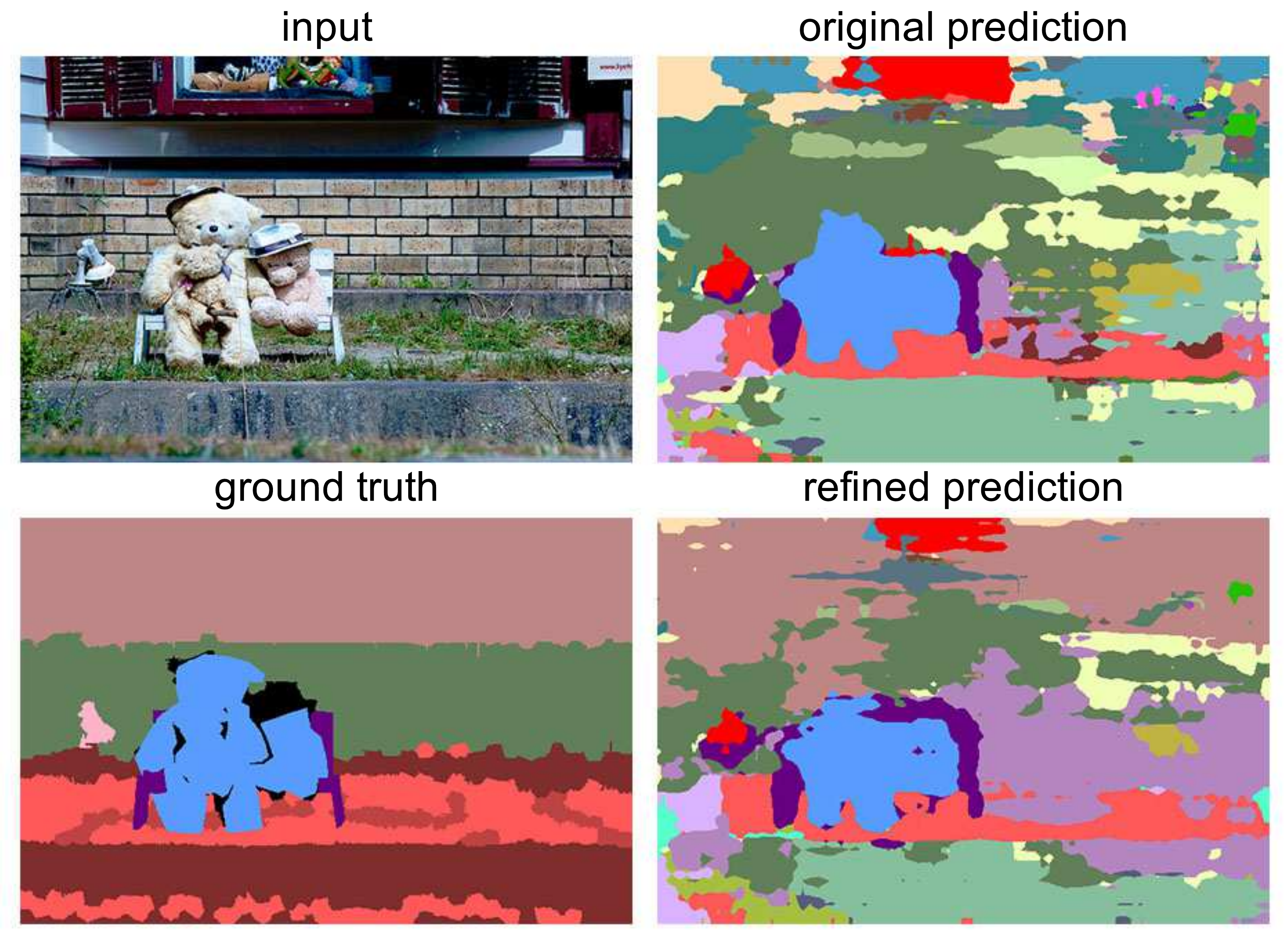}
	\caption{\textbf{Failure case.}. Hint: "\textit{there is a building in the top}" When the initial prediction fails, our method has difficulties recovering the mistakes. The refinement includes the \texttt{building} only partially and it bleeds into \texttt{stone-wall} below. } 
	\label{fig:badoripred}
\end{figure}

\vspace{-1em}
\paragraph{Failure Cases.}
When the initial prediction is particularly noisy, the guide has difficulties to fully repair the mistakes, as shown in Figure \ref{fig:badoripred}. Given a hint that a \texttt{building} is missing, the network can partly recover it, but a lot of spurious regions remain. We assume that the relevant features that would be needed for successful guiding, cannot be fully recovered from the noise in the guided activation map or are not present at all.

\begin{figure}[!t]
	\centering
	\includegraphics[clip,width=\linewidth]{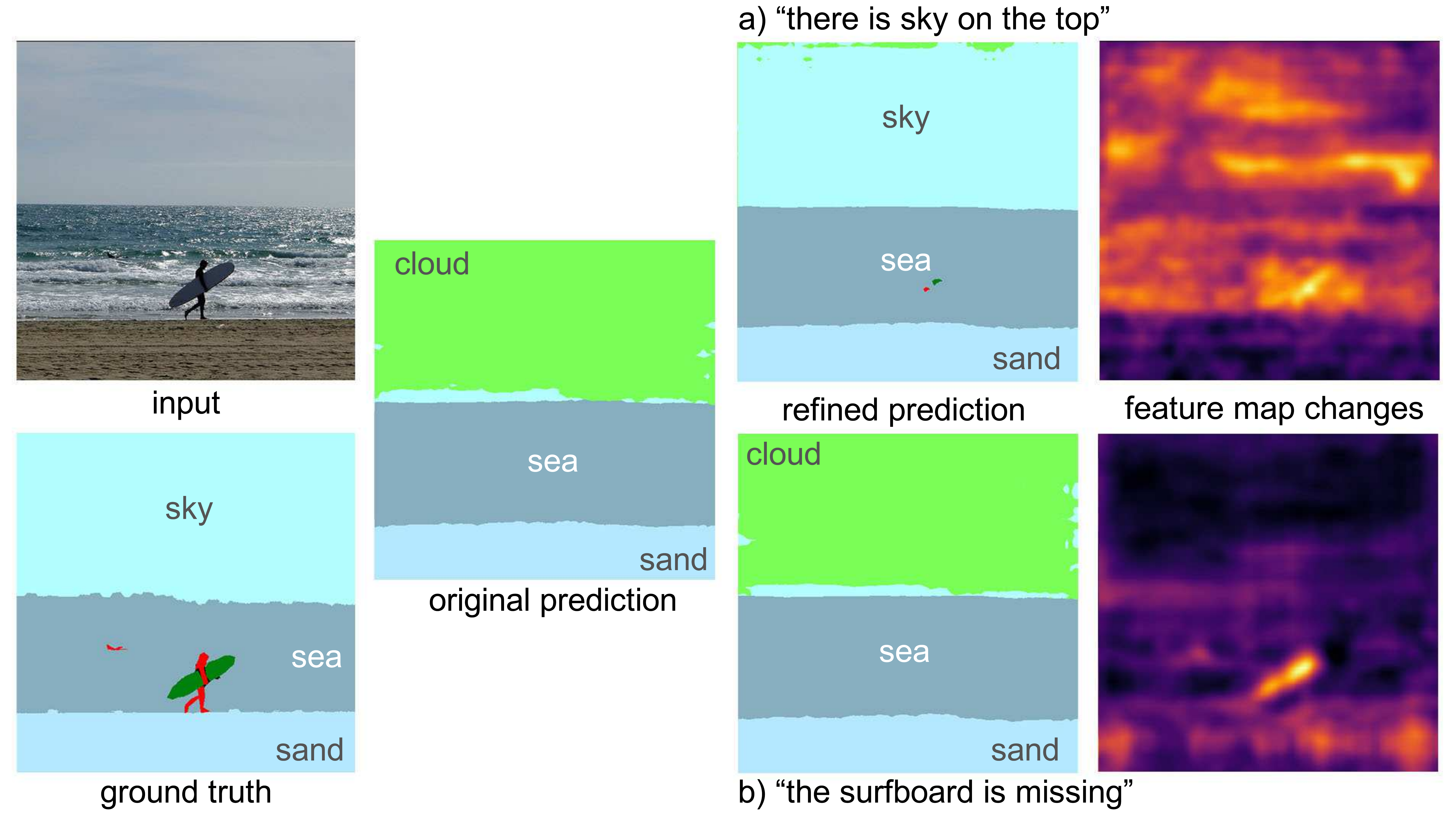}
	\caption{\textbf{Failure Case Visualization}. In the first example a) the refined prediction is correct. In b) the heatmap indicates that the guide has the right focus but it is not enough to change the output.} 
	\label{fig:heatmapfailure}
	\vspace{-0.5em}
\end{figure}

To understand how the activation map is influenced by the guide, we visualize a heatmap for different queries in the same image and investigate a failure case in Figure \ref{fig:heatmapfailure}. In this visualization we can see that the system understands the hint about the \texttt{sky} (a). However, given the refined prediction for the \texttt{surfboard} hint (b), we would assume that it did not understand the query correctly. The heatmap shows that the guide indeed does emphasize the right parts of the image, but not strong enough to overpower the \texttt{sea} label. 
Potentially more precise queries during training could fix this problem. "\textit{There is a surfboard where you predicted sea}" would let the guide not only emphasize \texttt{surfboard} related activations but simultaneously dampen the \texttt{sea} class, leading to better results in these cases.

\begin{figure}[t]
	\centering
	\includegraphics[trim = 55mm 50mm 40mm 20mm, clip,width=\linewidth]{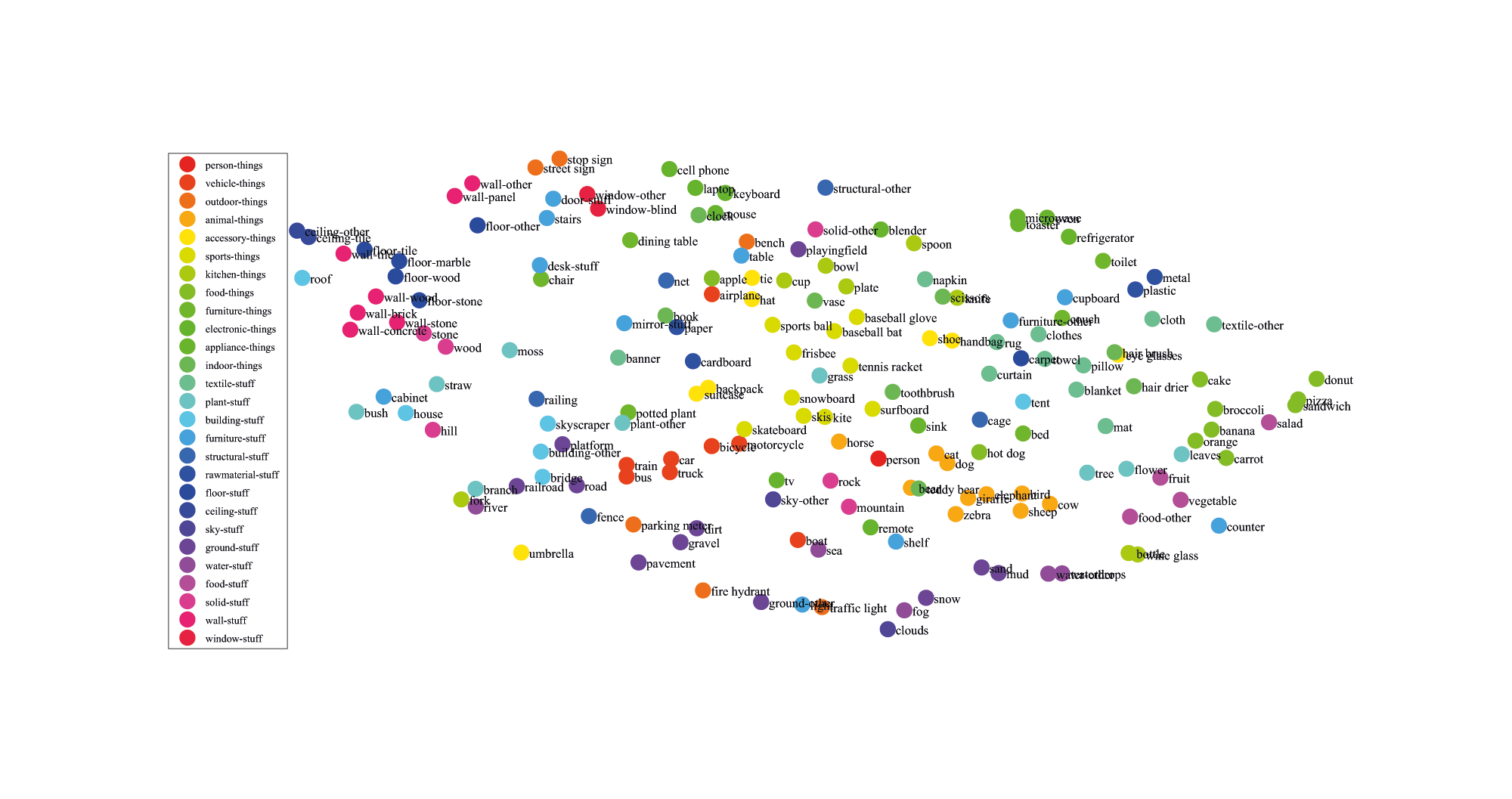}
	\caption{\textbf{Visualizing $\gamma$}. We visualize the learned $\gamma$ vector for every class using t-SNE. The colors correspond to the higher level categories which are present in the dataset but not used in training. Best viewed in digital version.} \label{fig:tsne}
\end{figure}

\vspace{-1em}
\paragraph{Semantic Analysis of the Learned $\gamma$-vectors.}
We analyze the mapping from text to guiding vectors. To this end, we predict a $\gamma$ vector for each class using a \texttt{find} query. In Figure \ref{fig:tsne} we display the t-SNE projection of these 256 dimensional vectors.
The color categories that the 182 classes are grouped into, are set from higher level categories. The grouping into categories was never used during training. 
This space is the intersection between features learned from the CNN for segmentation and text representation learned by the RNN. The fact that semantically similar words cluster means that the joint embedding successfully correlates text and image features.
A stronger clustering would mean that the $\gamma$-vectors are very similar inside the cluster, thus the network would have more difficulties guiding these classes. This can still be seen in a few cases such as the very close \textit{sand} and \textit{mud} classes, which are visually very similar and often do not improve after guiding.
\section{Conclusion}
In this paper, we have presented a system that allows for natural interaction of a human user with neural networks. The idea is to enable feedback from the user to guide the network by updating its feature representations on-the-fly, conditioned by the user's hint, without further training the network's parameters. An intuitive way of interaction is via text queries, sent by the human to the network, which aim at improving some initial estimation on a specific task. 

We have created queries automatically with a specialized algorithm. In the future we would like to explore the possibility of generating queries with a second network that learns the role of the user, giving hints to the first. Further, image-guided attention mechanism can be incorporated into the RNN to improve the interaction mechanism.

\section*{Acknowledgments}
We would like to thank Robert DiPietro for discussions about the idea and Helen L. Bear, Helisa Dhamo, Nicola Rieke, Oliver Scheel and Salvatore Virga for proofreading the manuscript and their valuable suggestions.

{\small
\bibliographystyle{ieee}
\bibliography{main}
}

\end{document}